\documentclass[11pt]{article}

\usepackage[preprint]{acl}

\usepackage{times}
\usepackage{latexsym}

\usepackage[T1]{fontenc}

\usepackage[utf8]{inputenc}

\usepackage{microtype}

\usepackage{inconsolata}

\usepackage{graphicx}
\usepackage{amsmath}
\usepackage{booktabs}
\usepackage{multirow}
\usepackage{amssymb}
\usepackage{enumitem}
\usepackage{graphbox}
\usepackage{changepage}
\usepackage[most]{tcolorbox}
\tcbuselibrary{breakable}
\usepackage[dvipsnames]{xcolor}
\usepackage{tikz}

\title{V-Zero: Self-Improving  Multimodal Reasoning with Zero Annotation}

\author{
 \textbf{Han Wang\textsuperscript{1*}},
 \textbf{Yi Yang\textsuperscript{1*}},
 \textbf{Jingyuan Hu\textsuperscript{2*}},
 \textbf{Minfeng Zhu\textsuperscript{2\textdagger}},
 \textbf{Wei Chen\textsuperscript{1\textdagger}}, 
\\
 \textsuperscript{1}State Key Laboratory of CAD\&CG, Zhejiang University,\\
 \textsuperscript{2}Zhejiang University
}

\begin{document}
\maketitle
\begin{tikzpicture}[remember picture, overlay]
    \node[anchor=north west, xshift=2.5cm, yshift=-1cm] 
        at (current page.north west) 
        {\includegraphics[width=6cm]{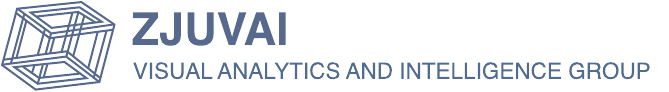}};
\end{tikzpicture}
\begingroup
  \renewcommand\thefootnote{} 
  \footnotetext{\textsuperscript{*}These authors contributed equally.}
  \footnotetext{\textsuperscript{\textdagger}Corresponding author}
\endgroup
\begin{figure*}[t!]
    \centering
    \includegraphics[height=6cm, align=c]{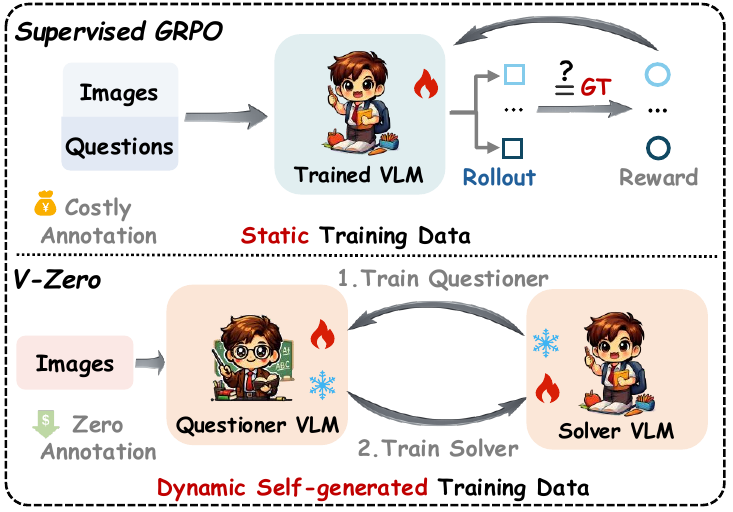} %
    \includegraphics[height=6cm, align=c]{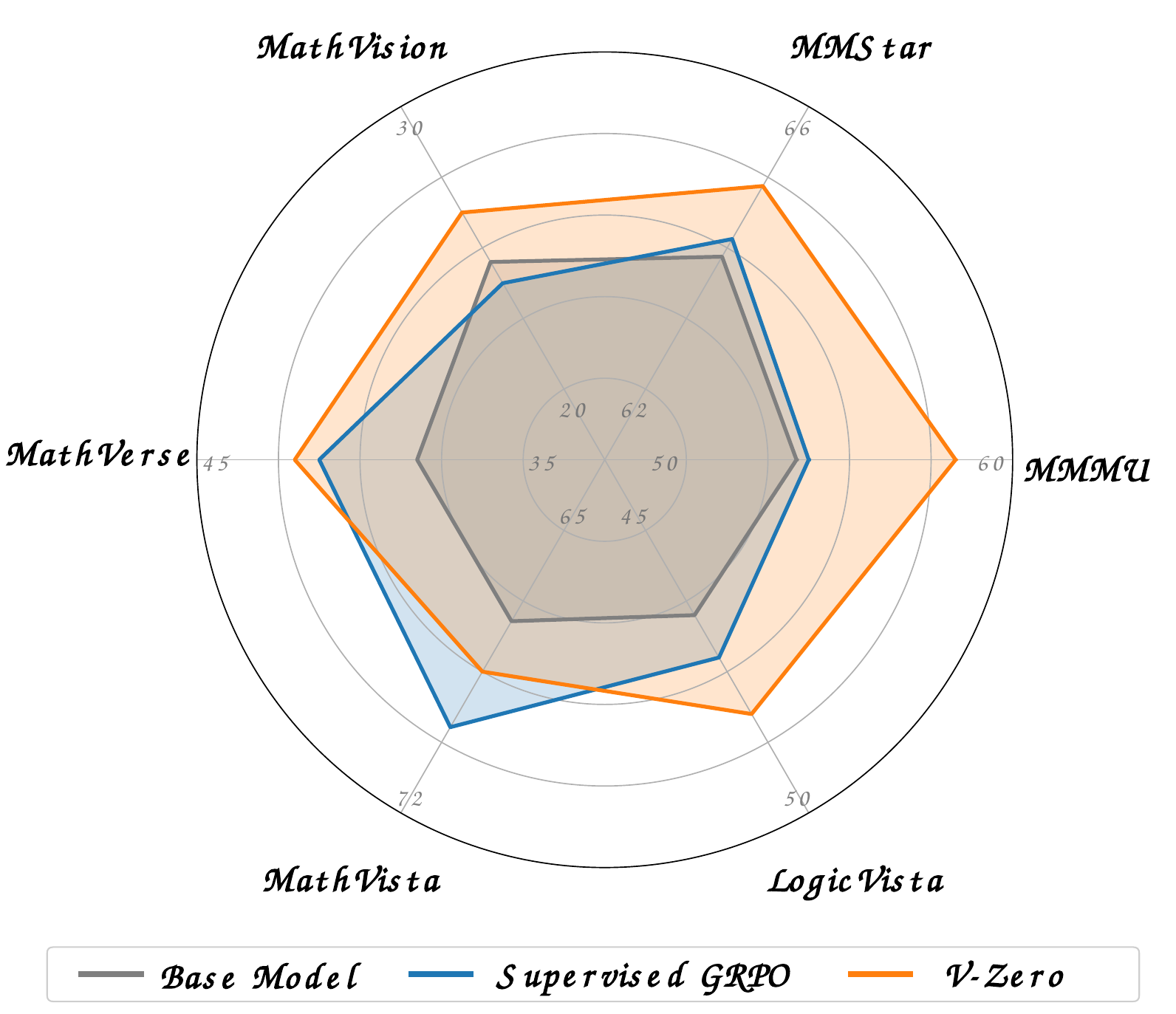}
    \caption{\textbf{Left:} Comparison of training paradigms. Supervised GRPO (top) relies on static datasets where rewards are computed by comparing outputs against external ground-truth labels. In contrast, V-Zero (bottom) achieves self-improvement from unlabeled images via a dynamic co-evolutionary loop, where reward signals are internally generated through the interaction between a Questioner and a Solver. \textbf{Right:} Performance on Qwen2.5-VL-7B-Instruct. V-Zero demonstrates broad improvements across diverse benchmarks, notably outperforming both the Base Model and the strong Supervised GRPO baseline trained on human-labeled data.}
    \label{fig:teaser}
\end{figure*}
\begin{abstract}

Recent advances in multimodal learning have significantly enhanced the reasoning capabilities of vision-language models (VLMs). However, state-of-the-art approaches rely heavily on large-scale human-annotated datasets, which are costly and time-consuming to acquire. To overcome this limitation, we introduce \textbf{V-Zero}, a general post-training framework that facilitates self-improvement using exclusively unlabeled images. V-Zero establishes a co-evolutionary loop by instantiating two distinct roles: a \textbf{Questioner} and a \textbf{Solver}. The Questioner learns to synthesize high-quality, challenging questions by leveraging a dual-track reasoning reward that contrasts intuitive guesses with reasoned results. The Solver is optimized using pseudo-labels derived from majority voting over its own sampled responses. Both roles are trained iteratively via Group Relative Policy Optimization (GRPO), driving a cycle of mutual enhancement. Remarkably, without a single human annotation, V-Zero achieves consistent performance gains on Qwen2.5-VL-7B-Instruct, improving visual mathematical reasoning by +1.7 and general vision-centric by +2.6, demonstrating the potential of self-improvement in multimodal systems.
\end{abstract}

\section{Introduction}
\label{sec:intro}

Vision-language models (VLMs) have become increasingly powerful at understanding images and solving complex reasoning problems. However, the performance of these models remains heavily dependent on extensive human annotation~\citep{bordes2024introduction}, such as visual question-answering (VQA) datasets containing fine-grained chain-of-thought (CoT)~\citep{wei2022chain} reasoning steps. Such high-quality reasoning annotations are not only costly and labor-intensive to acquire, but also typically designed for specific tasks or domains, resulting in limited generalization. In response, self-improvement methods have emerged as a compelling alternative, allowing models to iteratively enhance their capabilities using internal feedback rather than human guidance.

Self-improvement methods for large language models (LLMs) have demonstrated that models can autonomously generate training data and learning signals through self-play and role specialization interaction~\citep{kuba2025language, dong2025stp, liu2025spiral, huang2025r}. However, research on self-play for VLMs remains highly limited. Vision-Zero~\citep{wang2025vision} adapts the idea to VLMs by conducting self-play in games. However, it employs external expert models to edit and generate images during data construction. An important open question is whether VLM self-improvement can be achieved by operating directly on raw, unlabeled images without requiring annotation from humans or external models.  
\looseness=-1

To address this, we propose \textbf{V-Zero}, a general post-training framework that facilitates self-improvement using exclusively unlabeled images. 
As shown in the Figure~\ref{fig:teaser}, supervised Group Relative Policy Optimization (GRPO)~\citep{shao2024grpo} relies on static datasets consisting of human-annotated questions and ground-truth answers for supervised learning. 
However, V-Zero introduces a dynamic paradigm without annotated data by establishing a co-evolutionary loop between two roles initialized from the same base model: a Questioner and a Solver.
Both are trained with GRPO using internally generated reward signals. 
The Questioner learns to perceive an input image and generate multiple-choice questions that target the boundary of the Solver's current capabilities. Conversely, the Solver learns to answer these questions accurately. 
The Questioner maximizes a dual-track reasoning reward derived from the Solver's performance. By contrasting the model's initial intuition with its reasoned outputs, we encourage the Questioner to generate more challenging questions that cannot be solved by immediate guesses but require reflection and deep thinking. The Solver utilizes a binary accuracy reward within a reinforcement learning with verifiable rewards (RLVR) process~\citep{guo2025deepseekr1}, where pseudo-labels are determined via majority voting over sampled responses. Through this zero-annotation process, the Questioner and Solver progressively refine the model's visual reasoning faculties.
Our main contributions are as follows:
\begin{itemize}[leftmargin=*, nosep, align=left]
    \item We propose V-Zero, a general post-training framework that operates exclusively on image-only data without any annotation.
    \item We introduce a Dual-Track Reasoning Reward, which explicitly contrasts intuition and reasoning, to encourage the model to generate challenging, reasoning-intensive questions.
    \item Extensive experiments demonstrate that V-Zero substantially improves model performance across a range of tasks, even surpasses strong baselines trained with human-annotated supervision. 
\end{itemize}
\looseness=-1
\section{Related Work}

\subsection{Self-improvement of Large Models}
Self-improvement enhances model capabilities through iterative interactions in game-based settings and problem-solving tasks.

In game-based settings, models act as participants and achieve self-improvement by training in games and receiving feedback according to explicit rules. SPC~\citep{chen2025spc} frames this idea as an adversarial game between a sneaky generator injecting erroneous reasoning steps and a critic that must detect them, enabling step-level reliability learning without manual process annotation. SPIRAL~\citep{liu2025spiral} further scales competitive self-improvement to multi-turn, zero-sum games by using the same model to play both roles and optimizing it through role-conditioned advantages. In the multimodal field, Vision-Zero~\citep{wang2025vision} adapts self-improvement to VLMs by building a ``Who Is the Spy'' game on paired original and expert-edited images, where agents exchange clues and are trained with verifiable voting outcomes. 

In problem-solving scenarios, self-improvement follows a paired problem-generate-and-respond interaction. 
Absolute Zero~\citep{zhao2025absolute} realizes this interaction in coding-style reasoning by using a code executor to validate both generated problems and candidate solutions.
STP~\citep{dong2025stp} transfers the same principle to theorem proving by pairing a conjecturer with a prover, where conjectures are iteratively generated and filtered to form a self-curated curriculum. 
Beyond single-domain settings, SPICE~\citep{liu2025spice} broadens self-improvement to general tasks in corpus environments, where a challenger derives document-grounded questions and a reasoner answers them, producing an evolving curriculum through interactions. R-Zero~\citep{huang2025r} further generalizes this setup by jointly improving a challenger that pushes question difficulty and a solver that learns from its own sampled solutions.

However, current visual self-improvement remains largely constrained to the game-based scenario, which requires careful human design and time-consuming data preparation. To overcome these limitations, we propose a truly zero-human-involvement training scheme for VLMs that removes multimodal data preparation and runs a self-improving Questioner-Solver loop directly on raw, unlabeled images.

\subsection{Multimodal Reasoning}
Recent progress in advancing multimodal reasoning lies in exploring better reasoning trajectories and improving training signal~\citep{liu2024diving}.

Exploring better reasoning trajectories improves multimodal reasoning by retaining chains that best satisfy visual evidence and intermediate constraints~\cite{yao2024mulberry, sun2025mm}. MM-Verify~\citep{sun2025mm} builds an MM-Verifier to evaluate long CoT solutions, using verification to strengthen long-horizon multimodal reasoning, while GenPRM~\citep{zhao2025genprm} proposes a generative PRM that performs explicit reasoning to judge each step, enabling scalable test-time computation. Beyond verifier-guided selection, SoTA with Less~\citep{wang2025sota} uses MCTS-guided sample selection to prioritize high-leverage supervision for data-efficient self-improvement, and Mulberry~\citep{yao2024mulberry} extends collective MCTS to elicit reasoning and reflection trajectories with minimal reliance on humans. \looseness=-1

Improving training signals aims to strengthen multimodal reasoning by converting verifiable intermediate constraints into explicit rewards for optimization~\citep{deng2025openvlthinker, huang2025vision}. 
Vision-R1~\citep{huang2025vision} applies GRPO with a rule-based format and result rewards, and leverages a cold-start multimodal CoT initialization to make such simple signals effective. 
R1-Onevision~\citep{yang2025r1} combines rule-based answer checking with strict reasoning-format constraints, thereby enhancing complex problem-solving. 
To address the sparsity of outcome-only supervision, R1-VL~\citep{zhang2025r1} introduces dense step-wise rewards that credit matched key intermediate steps and encourage complete, logically consistent trajectories. 
Beyond language-centric reasoning rewards, Perception-R1~\citep{yu2025perception} designs task-specific rewards for perception outputs and performs multi-subject reward matching before scoring, providing reliable training signals for learning perception policies.

Compared with these approaches relying on large-scale annotated data, our method eliminates data annotation by training directly on unlabeled images, yet outperforms supervised training. 
\section{Method}
\label{sec:method}

\begin{figure*}[t!]
    \centering
    \includegraphics[width=\textwidth]{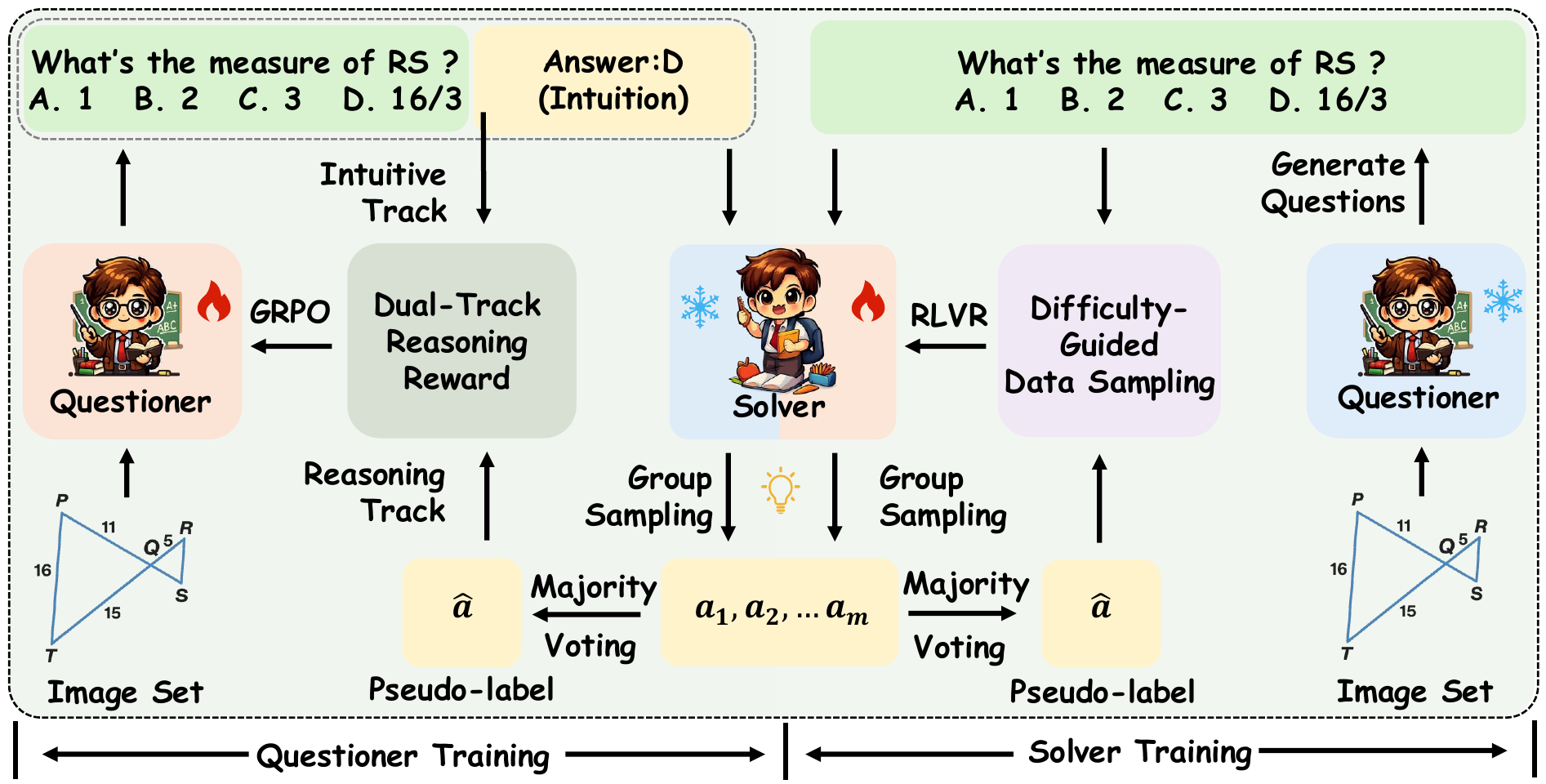} 
    \caption{\textbf{Overview of V-Zero.} Our framework drives self-improvement through a co-evolutionary loop. \textbf{Left (Questioner Training):} The Questioner generates a question and an intuitive answer. A frozen Solver then samples reasoning responses to produce a pseudo-label via majority voting. The Questioner is updated via GRPO using a Dual-Track Reasoning Reward derived from the contrast between intuition and reasoning. \textbf{Right (Solver Training):} The optimized Questioner generates questions, which are paired with pseudo-labels from the Solver itself. This data undergoes difficulty-guided data sampling to filter for quality, and the Solver is then updated via RLVR.}
    \label{fig:pipeline}
\end{figure*}

\subsection{Overview}
We propose \textbf{V-Zero}, a general post-training framework for VLMs that operates exclusively on raw image data without any annotations, as illustrated in Figure~\ref{fig:pipeline}. The framework establishes a co-evolutionary loop between two roles initialized from the same base model: a Questioner $Q_\theta$ and a Solver $S_\theta$. The Questioner perceives input images to generate challenging, high-quality questions, optimized via GRPO using feedback signals from the Solver. Correspondingly, the Solver learns to answer these questions via GRPO using a difficulty-guided sampled dataset.
Through iterative cycles, $Q_\theta$ and $S_\theta$ continuously co-evolve in a self-supervised manner, progressively enhancing visual perception and reasoning capabilities.
\looseness=-1

\subsection{Questioner}
\label{method:questioner}
Questioner model $Q_\theta$ learns to understand input images from an image-only dataset and generate high-quality questions that are maximally challenging for the Solver. 

\paragraph{Image-based Question Generation}
Given an image input $I$, $Q_\theta$ fully perceives the image and extracts useful visual information, such as side lengths of geometric shapes, entries in charts, spatial relationships between objects, and other salient visual features. $Q_\theta$ then generates a concise visual description $d$ of $I$, a multiple-choice question (MCQ) $q$, and a reference answer $a_{fast}$ indicating the correct option for the question.
\begin{equation}
    (q,a_{fast},d) \sim Q_\theta(\cdot|I)
\end{equation}
Here, $a_{fast}$ represents the model's "direct answer" or intuition, as it is generated simultaneously with the question without extensive thinking. We specifically adopt the MCQ format for its structural advantages: it constrains the output to a fixed set of options, facilitating robust answer extraction and enabling precise, automated evaluation of the Solver's accuracy without the ambiguity of free-form generation.


\paragraph{Dual-Track Reasoning Reward}
A widely accepted principle is that, for easy problems, answers derived from intuition and careful reasoning tend to coincide; in contrast, a defining characteristic of challenging problems is that intuitive responses are often misleading, whereas correct solutions can be reached through step-by-step reasoning. Inspired by this observation, we implement a dual-track reasoning reward to encourage the Questioner $Q_\theta$ to generate such questions. We contrast two distinct answering modes:
\begin{itemize}[leftmargin=*, nosep, align=left]
    \item \textbf{Intuitive Track ($a_{fast}$):} The direct answer generated by the Questioner simultaneously with the question, representing the model's intuition.
    \item \textbf{Reasoning Track ($\hat{a}$):} The pseudo-label derived from the Solver $S_\theta$. The Solver employs CoT reasoning to sample $m$ responses $\{a_1, \dots, a_m\}$. The majority vote determines $\hat{a}$, with confidence $c$ defined as the proportion of votes matching $\hat{a}$.
\end{itemize}
The dual-track reasoning reward $r_{d}$ is adaptive based on the alignment between the intuitive track and the reasoning track:
\begin{equation}
    r_{d}(q) = 
\begin{cases} 
    \min(c, 1-c), & \text{if } \hat{a} = a_{fast}, \\
    0.5 \cdot c, & \text{otherwise}.
\end{cases}
\end{equation}
\begin{itemize}[leftmargin=*, nosep, align=left]
    \item \textbf{Consistency Case ($\hat{a} = a_{fast}$):} When intuition aligns with reasoning, the problem is likely straightforward. In this case, we adopt an uncertainty reward. If the confidence $c$ is high (near 1.0), it implies the question is overly simple; if $c$ is low (near 0), it implies the question is overly ambiguous or noisy. The reward $\min(c, 1-c)$ peaks at $c=0.5$, effectively guiding the Questioner to locate the Solver's decision boundary, targeting questions that are neither too simple nor completely ambiguous but maximally confusing for the current model.
    \item \textbf{Divergence Case ($\hat{a} \neq a_{fast}$):} When the reasoned answer corrects the intuitive guess,  it implies the question serves as a valuable learning signal. Here, the reward is proportional to the confidence $c$, encouraging the generation of challenging questions that appear simple at first glance but require the Solver to engage in reflection and deep reasoning to obtain the correct answer with high confidence. \looseness=-1
    \item \textbf{Prevention of Reward Hacking:} Our design naturally prevents reward hacking. Nonsensical questions lead to inconsistent reasoning (low $c$), yielding low reward in divergence case. Conversely, generating trivial questions to boost $c$ causes answers to converge ($\hat{a} = a_{fast}$), triggering the uncertainty reward where high confidence yields near-zero reward. Thus, high rewards are strictly reserved for valid, challenging questions.
\end{itemize}

\paragraph{Final Reward with Formatting Constraints}
To ensure structural validity, we enforce strict formatting constraints. A generation pair $(q, a_{fast})$ is valid only if enclosed within distinct \textit{<question>} and \textit{<answer>} tags, and $q$ is formatted as a multiple-choice question with four options. The final reward for the Questioner is:
\begin{equation}
    r_Q = 
\begin{cases} 
    r_{d}(q), & \text{if format is valid}, \\
    -1, & \text{otherwise}.
\end{cases}
\end{equation}

\subsection{Solver}
Solver model $S_\theta$ learns to answer the questions proposed by $Q_\theta$. In this phase, we focus on curating a high-quality training dataset through difficulty-guided data sampling and optimizing the model's reasoning capabilities via RLVR.\looseness=-1

\paragraph{Difficulty-Guided Data Sampling}
Given a set of $n$ unlabeled images $\{I_i\}_{i=1}^n$, the Questioner $Q_\theta$ first generates a corresponding multiple-choice question $q_i$ for each image. The Solver $S_\theta$ then samples a set of $m$ candidate answers:
\begin{equation}
    \{a_{i,1}, a_{i,2}, \dots, a_{i,m}\} \sim S_\theta(\cdot \mid (I_i, q_i))
\end{equation}
From these samples, a pseudo-label $\hat{a}_i$ is derived via majority voting, with the vote proportion recorded as a difficulty score $s_i$. To curate a training set that effectively drives improvement, we implement a difficulty-guided sampling strategy, keeping only pairs $(q_i, \hat{a}_i)$ where the score falls between $0.3 \leq s_i \leq 0.8$. This rule filters out samples that are either too ambiguous (low $s_i$, suggesting potential errors) or too simple (high $s_i$, offering little learning value). The final dataset $D$ is thus constructed by selecting questions that are challenging yet solvable for the current model:
\begin{equation}
\begin{aligned}
D = \bigl\{ (I_i, q_i, \hat{a}_i, s_i) \;\big|\;&\; i \in \{1,\dots,n\}, \\
&\; 0.3 \leq s_i \leq 0.8 \bigr\}
\end{aligned}
\end{equation}

\paragraph{Binary Accuracy Reward}
The Solver $S_\theta$ is optimized to maximize answer accuracy on the constructed dataset $D$. Specifically, for each question $q_i$ paired with its pseudo-label $\hat{a}_i$, the Solver samples a new group of candidate answers $\{a_{i,j}\}_{j=1}^G$. The reward $r_j$ assigned to each candidate is defined as a binary value: 
\begin{equation}
    r_j = 
\begin{cases} 
    1, & \text{if } a_{i,j} = \hat{a}_i, \\
    0, & \text{otherwise}.
\end{cases}
\end{equation}
The binary accuracy rewards are subsequently utilized to compute the group relative advantage $\hat{A}_{i,j}$. Specifically, the rewards within each group are standardized using z-score normalization:
\begin{equation}
    \hat{A}_{i,j} = \frac{r_j - \text{mean}(\{r_1, \dots, r_G\})}{\text{std}(\{r_1, \dots, r_G\}) + \epsilon_{\text{norm}}}.
\end{equation}
Finally, the policy is updated by minimizing the GRPO loss, which combines a clipped surrogate objective with a KL-divergence penalty:
\begin{equation}
\begin{split}
    \mathcal{L}_{\text{GRPO}}(\theta) = 
    & -\frac{1}{G} \sum_{j=1}^G \min \bigg( \frac{\pi_\theta(a_{i,j})}{\pi_{\text{old}}(a_{i,j})} \hat{A}_{i,j}, \\
    & \text{clip}\Big(\frac{\pi_\theta(a_{i,j})}{\pi_{\text{old}}(a_{i,j})}, 1-\epsilon, 1+\epsilon\Big) \hat{A}_{i,j} \bigg) \\
    & + \beta \text{KL}(\pi_\theta || \pi_{\text{old}}).
\end{split}
\end{equation}

\section{Experiments}

\begin{table*}[t]
    \centering
    \small
    \setlength{\tabcolsep}{6pt} 
    \begin{tabular}{l|ccccccc}
    \toprule
    \multirow{2}{*}{\textbf{Methods}} & \multicolumn{2}{c}{\textbf{General Vision-Centric}} & \multicolumn{4}{c}{\textbf{Mathematical Reasoning}} & \multirow{2}{*}{\textbf{Avg.}} \\
    \cmidrule(lr){2-3} \cmidrule(lr){4-7}
    & \textbf{MMMU} & \textbf{MMStar} & \textbf{MathVision} & \textbf{MathVerse} & \textbf{MathVista} & \textbf{LogicVista} & \\
    \midrule
    \multicolumn{8}{c}{\textit{Performance on Qwen2.5-VL-7B-Instruct}} \\
    \midrule
    Base Model & 54.7 & 63.9 & 25.6 & 39.6 & 68.2 & 47.2 & 49.9 \\
    Supervised GRPO & 55.0 & 64.5 & 25.0 & 42.0 & 70.3 & 47.8 & 50.8 \\
    OpenVLThinker-7B & 53.1 & 59.6 & \textbf{27.6} & 36.8 & \textbf{70.8} & 45.1 & 48.8 \\
    V-Zero (Iter 1) & 55.2 & \textbf{65.7} & 26.3 & \textbf{43.9} & 68.6 & 47.4 & 51.2 \\
    V-Zero (Iter 2) & \textbf{58.6} & 65.1 & 27.0 & 42.6 & 69.2 & \textbf{48.6} & \textbf{51.9} \\
    \midrule
    \multicolumn{8}{c}{\textit{Performance on Qwen2.5-VL-3B-Instruct}} \\
    \midrule
    Base Model & 45.6 & 56.1 & 23.0 & 32.1 & 62.2 & 40.4 & 43.2 \\
    V-Zero (Iter 1) & \textbf{47.3} & 54.7 & \textbf{25.3} & \textbf{33.4} & \textbf{62.6} & 40.2 & \textbf{43.9} \\
    V-Zero (Iter 2) & 46.7 & \textbf{56.4} & 23.7 & 32.6 & 62.0 & \textbf{41.6} & 43.8 \\
    \bottomrule
    \end{tabular}
    \caption{Performance comparison on general vision-centric and mathematical reasoning benchmarks, evaluated on VLMEvalKit. All results are obtained under same settings. We report results of Solver for both Qwen2.5-VL-7B-Instruct and Qwen2.5-VL-3B-Instruct. Best results are highlighted in \textbf{bold}. }
    \label{tab:main_results}
    \vspace{-10pt}
\end{table*}

\subsection{Settings}
\paragraph{Models \& Datasets} 
We conduct experiments on two state-of-the-art VLM backbones: the widely adopted Qwen2.5-VL-3B-Instruct and Qwen2.5-VL-7B-Instruct~\citep{bai2025qwen2_5vl}. For training data, we use images from the OpenVLThinker GRPO-medium and GRPO-hard dataset~\citep{deng2025openvlthinker}, comprising approximately 9K images covering multiple task types, primarily geometry with around 6K images, tables and spatial reasoning with around 3K images in total. After an initial filtering of the dataset, we train V-Zero exclusively on the image data. More details about image dataset can be found in Appendix~\ref{sec:appendix:dataset}.

\paragraph{Baselines}
We compare our method against the base model and a model trained with two epochs of supervised GRPO on the same image dataset but fully human-annotated. We also include OpenVLThinker-7B as a stronger baseline, which is post-trained via supervised fine-tuning (SFT) and GRPO using human-annotated data.

\paragraph{Evaluation Benchmarks}
We evaluate V-Zero using VLMEvalKit~\citep{duan2024vlmevalkit} across benchmarks in two multimodal domains: general vision-centric tasks and visual mathematical reasoning. For general vision-centric evaluation, we employ MMMU~\citep{yue2024mmmu} and MMStar~\citep{chen2024mmstar}. MMMU evaluates college-level subject knowledge and reasoning with 11.5K multimodal questions across six disciplines. MMStar tests core VLM capabilities on vision-indispensable tasks through 1,500 meticulously curated samples. For mathematical reasoning, we employ MathVision~\citep{wang2024mathvision}, MathVerse~\citep{zhang2024mathverse}, MathVista~\citep{lu2024mathvista} and LogicVista~\citep{xiao2024logicvista}. MathVision is a comprehensive benchmark comprising 3,040 high-quality visual mathematics problems sourced from real competitions. We use the Test-mini split consisting of 304 samples. MathVerse is a robust visual math benchmark containing 2,612 multi-subject problems. We report results on the Vision-Only split, as it presents the most significant challenge for the models. MathVista integrates 6,141 examples from 31 diverse datasets to evaluate fine-grained visual understanding and compositional reasoning in mathematical contexts. We use the Test-mini split, consisting of approximately 1,000 samples. LogicVista evaluates fundamental logical reasoning in visual contexts through 448 multiple-choice questions.\looseness=-1

\paragraph{Training Settings}
Our implementation is based on the verl framework~\citep{sheng2024hybridflow}. Both the Questioner and Solver are optimized via GRPO on 4 NVIDIA A800 (80GB) GPUs with a global batch size of 64, a learning rate of $1\times10^{-6}$, and a sampling temperature of 1.0. For the Questioner, we dedicate 2 additional GPUs to host the feedback Solver and set the group size $G=4$. For Solver training, we set $G=5$. Additionally, we set the sample size for majority voting to $m=10$. More details about training settings can be found in Appendix~\ref{sec:appendix:training}.
\subsection{Main Results}

\paragraph{Overall Performance Improvements.} 
Table \ref{tab:main_results} presents the evaluation results of the Solver across diverse benchmarks. V-Zero consistently enhances the performance of the base models without external supervision. For Qwen2.5-VL-7B-Instruct, the iterative training yields a steady performance trajectory for the Solver, improving the average score from 49.9 to 51.9 (+2.0 points) in Iteration 2. The gains are particularly substantial on the general benchmark MMMU (+3.9 points) and the visual mathematics dataset MathVerse (+3.0 points), validating the framework's effectiveness. For the smaller Qwen2.5-VL-3B-Instruct, the model reaches its performance peak in the first iteration (+0.7 points), with notable gains on MMMU (+1.7 points) and MathVision (+2.3 points). We attribute this early saturation to the restricted model capacity and capability, which increases sensitivity to the noise in self-generated data. Furthermore, the requisite high-temperature sampling for exploration inherently introduces optimization variance, a trade-off that becomes more pronounced in smaller models with limited robustness.

\paragraph{Generalization Across Tasks.} Our training image dataset is primarily composed of geometric figures aimed at enhancing mathematical reasoning capabilities, validated by the consistent improvements observed on mathematical reasoning benchmarks (+1.7 points on average). Importantly, these specialized gains do not compromise general capabilities. Instead, V-Zero demonstrates strong generalization, achieving even larger improvements on general vision-centric datasets (+2.6 points on average). This suggests that the rigorous visual perception required to solve complex geometric problems effectively transfers to broader domains, strengthening the Solver's fundamental visual understanding without overfitting to a specific domain.

\paragraph{Comparison with Supervised Baselines.} V-Zero compares favorably against methods relying on ground-truth supervision. On the 7B scale, our unsupervised Solver (51.9 avg.) outperforms the Supervised GRPO baseline (50.8 avg.), which was trained using human-annotated questions and answers. This indicates that the diversity of questions generated by the Questioner may offer better exploration of the solution space than a fixed supervised dataset. Furthermore, compared to OpenVLThinker-7B, V-Zero remains highly competitive. While OpenVLThinker holds a marginal advantage in specific mathematical reasoning benchmarks like MathVista and MathVision, V-Zero achieves superior performance on LogicVista and maintains a more balanced profile across general vision-centric tasks.

\subsection{Ablation Studies}

To verify the effectiveness of the core components in V-Zero, we conduct comprehensive ablation studies on the Qwen2.5-VL-7B-Instruct model. We compare the full V-Zero (Iteration 2) against three variants: freezing the Questioner, removing the dual-track reasoning reward, and disabling difficulty-guided data sampling for Solver training. The results are summarized in Table \ref{tab:ablation}.

\paragraph{Impact of Improving Questioner}
We first investigate the validity of the framework by freezing the Questioner at its initial state (base model) while only training the Solver. As shown in Table \ref{tab:ablation}, this setting leads to a catastrophic performance drop. The Math Avg. falls to 45.4, showing negligible improvement over the Base Model (45.2). More critically, the General Avg. degrades to 58.7, which is even lower than the Base Model (59.3). This suggests that without the improvement of the Questioner, the generated questions remain static and lack sufficient challenge. Consequently, the Solver likely overfits to simple patterns or suffers from distribution shift, ultimately harming its general visual capabilities. This confirms that the dynamic co-evolution between the two roles is essential for continuous self-improvement.

\begin{table}[t]
    \centering
    \small
    \setlength{\tabcolsep}{3pt} 
    \begin{tabular}{lcc}
    \toprule
    \textbf{Methods} & \textbf{General Avg.} & \textbf{Math Avg.}\\
    \midrule
    Base Model & 59.3 & 45.2\\
    \textbf{V-Zero (Iter 2)} & \textbf{61.9} & \textbf{46.9}\\
    \midrule
    \quad w/o. Improving Questioner & 58.7 & 45.4\\
    \quad w/o. Dual-Track Reward & 61.3 & 46.1 \\
    \quad w/o. Data Filtering & 59.8  & 45.8\\
    \bottomrule
    \end{tabular}
    \caption{Ablation studies on key components of V-Zero. We evaluate the impact of the improving Questioner, dual-track reasoning reward, and difficulty-guided data sampling (denoted as Data Filtering) on Qwen2.5-VL-7B-Instruct. Best results are highlighted in \textbf{bold}.}
    \label{tab:ablation}
\end{table}
\paragraph{Questioner Reward Design}
To assess the contribution of our proposed Dual-Track Reasoning Reward, we replace it with a single uncertainty reward (i.e., removing the reward of divergence case). The ablation results show a clear performance drop, with the Math Avg. dropping from 46.9 to 46.1 (-0.8 points) and the General Avg. decreasing from 61.9 to 61.3 (-0.6 points). This decline indicates that simply targeting high-uncertainty samples is insufficient. By explicitly rewarding questions where the model’s intuition contrasts with its reasoning, the dual-track reasoning reward successfully encourages the Questioner to generate more challenging questions, thereby forcing the Solver to engage in deeper reasoning.

\paragraph{Impact of Data Filtering}
Finally, we examine the role of the difficulty-guided data sampling strategy (selecting samples with $0.3 \le s \le 0.8$). Removing this filter results in a significant performance decline, with the General Avg. dropping by 2.1 points and Math Avg. dropping by 1.1 points. This underscores the importance of data quality in the self-improvement process. Without data filtering, Solver training may be contaminated by two types of data: overly simple samples that contribute nothing to performance improvement, and noisy samples arising from ambiguous or unsolvable problems. The filtering mechanism ensures that the Solver learns exclusively from helpful data, thereby preserving training stability and efficiency.

\subsection{Qualitative Analysis}
\begin{figure*}[t]
    \centering
    \includegraphics[width=0.9\textwidth]{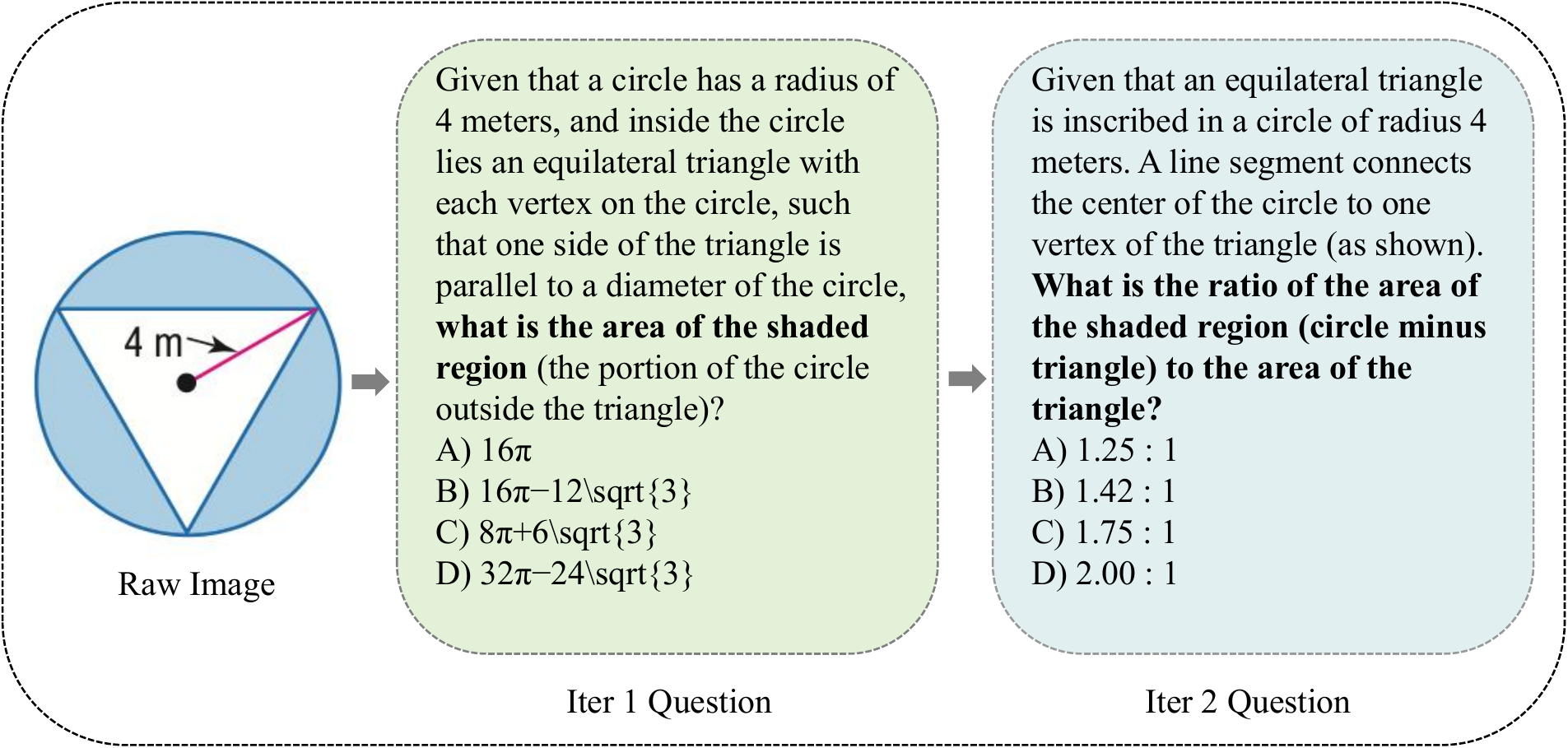} 
    \caption{\textbf{Improvement of generated questions.} We visualize the questions generated by the Questioner for the same raw image (Left) across two iterations. In Iteration 1 (Middle), the model poses a straightforward area calculation problem. In Iteration 2 (Right), the question advances into a more complex ratio problem involving deeper geometric reasoning, demonstrating the progressive difficulty of the self-generated curriculum.}
    \label{fig:question}
\end{figure*}

We analyze the quality of the questions generated by the Questioner in the training process, focusing on two key metrics: format validity and question difficulty. The statistics are summarized in Table \ref{tab:valid-question}.
\paragraph{Format Validity}

As described in Section \ref{method:questioner}, we impose strict formatting constraints during Questioner training: a generated question sample is considered valid only if it presents a four-option multiple-choice question with correctly enclosed tags for the question and the intuitive answer. As shown in Table \ref{tab:valid-question}, the Base Model struggles with these structural requirements, achieving a validity rate of only 64.9\%, which severely limits the utilization of generated data. However, after just one iteration of V-Zero training, the validity rate surges to 99.1\%. This near-perfect compliance demonstrates that the constraints effectively regularizes the model's output format, ensuring the production of high-quality, parsable training data.
\begin{table}[t]
    \centering
    \setlength{\tabcolsep}{8pt}
    \begin{tabular}{lcc}
    \toprule
    \textbf{Questioner} & \textbf{Valid Rate} & \textbf{Difficulty}\\
    \midrule
    Base Model & 64.9\% & 0.52\\
    V-Zero (Iter 1) & 99.1\% & 0.58\\
    V-Zero (Iter 2) & \textbf{99.8\%} & \textbf{0.60}\\
    \bottomrule
    \end{tabular}
    \caption{Evolution of generated question quality. We report the percentage of valid questions and the average difficulty  evaluated by Qwen3-VL-32B-Instruct. Best results are highlighted in \textbf{bold}.}
    \label{tab:valid-question}
\end{table}
\paragraph{Question Difficulty}
To evaluate the depth of the generated questions, we employ the advanced Qwen3-VL-32B-Instruct~\cite{bai2025qwen3vl} as a judge. We establish a scoring rule ranging from 0 to 1, categorizing questions into three levels:
\begin{itemize}[nosep]
    \item Simple (0.0-0.3): Basic object counting or simple chart reading.
    \item Medium (0.4-0.7): Application of geometric relations or chart trend analysis.
    \item Hard (0.8-1.0): Complex geometric proofs or multi-step algebraic reasoning.
\end{itemize}
We calculate the average difficulty score for all valid questions. The quantitative results indicate a clear upward trend, rising from 0.52 (Base) to 0.60 (Iter 2). This progression is shown in Figure \ref{fig:question}: while the model in Iteration 1 poses a straightforward area calculation problem, it evolves in Iteration 2 to construct a significantly more complex ratio problem requiring multi-step derivation. This confirms that the Questioner learns to generate harder questions over time, thereby providing a curriculum of growing difficulty that continuously challenges the Solver.

\section{Conclusion}
In this work, we introduced V-Zero, a general framework enabling the self-improvement of VLMs utilizing exclusively unlabeled images. By establishing a dynamic co-evolutionary loop between a Questioner and a Solver, V-Zero effectively bypasses the bottleneck of costly human annotations. In this process, the Questioner learns to generate increasingly challenging questions by targeting the gap between intuition and reasoning, while the Solver iteratively improves by resolving these generated questions. Both roles are optimized via GRPO. Extensive experiments demonstrate that this paradigm not only yields significant gains in visual mathematical reasoning but also generalizes well to broader vision-centric tasks. These results suggest that constructing general, internal feedback loops is a viable path toward more general multimodal capabilities without reliance on costly human supervision.

\section{Limitations}
First, due to compute constraints, our experiments are limited to the Qwen2.5-VL series. Our future work will extend V-Zero to more model architectures.
Second, experimental results indicate that the range of performance improvement is constrained by the model’s intrinsic capacity, and that the optimization process exhibits fluctuations rather than steady growth across iterations. These fluctuations are partly due to the requirement to encourage high-degree exploration for unlocking the model’s latent potential.



\bibliography{custom}

@String(ICLR = {Int. Conf. Learn. Represent.})

@String(ICLR  = {ICLR})

@article{bordes2024introduction,
  title={An introduction to vision-language modeling},
  author={Bordes, Florian and Pang, Richard Yuanzhe and Ajay, Anurag and Li, Alexander C and Bardes, Adrien and Petryk, Suzanne and Ma{\~n}as, Oscar and Lin, Zhiqiu and Mahmoud, Anas and Jayaraman, Bargav and others},
  journal={arXiv preprint arXiv:2405.17247},
  year={2024},
  url={https://arxiv.org/abs/2405.17247}
}

@article{shao2024grpo,
  title={Deepseekmath: Pushing the limits of mathematical reasoning in open language models},
  author={Shao, Zhihong and Wang, Peiyi and Zhu, Qihao and Xu, Runxin and Song, Junxiao and Bi, Xiao and Zhang, Haowei and Zhang, Mingchuan and Li, YK and Wu, Yang and others},
  journal={arXiv preprint arXiv:2402.03300},
  year={2024},
  url={https://arxiv.org/abs/2402.03300}
}

@inproceedings{wei2022chain,
  author       = {Jason Wei and
                  Xuezhi Wang and
                  Dale Schuurmans and
                  Maarten Bosma and
                  Brian Ichter and
                  Fei Xia and
                  Ed H. Chi and
                  Quoc V. Le and
                  Denny Zhou},
  editor       = {Sanmi Koyejo and
                  S. Mohamed and
                  A. Agarwal and
                  Danielle Belgrave and
                  K. Cho and
                  A. Oh},
  title        = {Chain-of-Thought Prompting Elicits Reasoning in Large Language Models},
  booktitle    = {Advances in Neural Information Processing Systems 35: Annual Conference
                  on Neural Information Processing Systems 2022, NeurIPS 2022, New Orleans,
                  LA, USA, November 28 - December 9, 2022},
  year         = {2022},
  url          = {https://arxiv.org/abs/2201.11903},
  bibsource    = {dblp computer science bibliography, https://dblp.org}
}

@article{guo2025deepseekr1,
  title={Deepseek-r1: Incentivizing reasoning capability in llms via reinforcement learning},
  author={Guo, Daya and Yang, Dejian and Zhang, Haowei and Song, Junxiao and Zhang, Ruoyu and Xu, Runxin and Zhu, Qihao and Ma, Shirong and Wang, Peiyi and Bi, Xiao and others},
  journal={arXiv preprint arXiv:2501.12948},
  year={2025},
  url={https://arxiv.org/abs/2501.12948}
}

@article{bai2025qwen2_5vl,
  title={Qwen2.5-VL Technical Report},
  author={Bai, Shuai and Chen, Keqin and Liu, Xuejing and Wang, Jialin and Ge, Wenbin and Song, Sibo and Dang, Kai and Wang, Peng and Wang, Shijie and Tang, Jun and others},
  journal={arXiv:2502.13923},
  year={2025},
  url={https://arxiv.org/abs/2502.13923}
}

@article{bai2025qwen3vl,
  title={Qwen3-VL Technical Report},
  author={Shuai Bai and Yuxuan Cai and Ruizhe Chen and Keqin Chen and Xionghui Chen and Zesen Cheng and Lianghao Deng and Wei Ding and Chang Gao and Chunjiang Ge and Wenbin Ge and Zhifang Guo and Qidong Huang and Jie Huang and Fei Huang and Binyuan Hui and Shutong Jiang and Zhaohai Li and Mingsheng Li and Mei Li and Kaixin Li and Zicheng Lin and Junyang Lin and Xuejing Liu and Jiawei Liu and Chenglong Liu and Yang Liu and Dayiheng Liu and Shixuan Liu and Dunjie Lu and Ruilin Luo and Chenxu Lv and Rui Men and Lingchen Meng and Xuancheng Ren and Xingzhang Ren and Sibo Song and Yuchong Sun and Jun Tang and Jianhong Tu and Jianqiang Wan and Peng Wang and Pengfei Wang and Qiuyue Wang and Yuxuan Wang and Tianbao Xie and Yiheng Xu and Haiyang Xu and Jin Xu and Zhibo Yang and Mingkun Yang and Jianxin Yang and An Yang and Bowen Yu and Fei Zhang and Hang Zhang and Xi Zhang and Bo Zheng and Humen Zhong and Jingren Zhou and Fan Zhou and Jing Zhou and Yuanzhi Zhu and Ke Zhu},
  journal={arXiv preprint arXiv:2511.21631},
  year={2025},
  url={https://arxiv.org/abs/2511.21631}
}

@inproceedings{deng2025openvlthinker,
  title={Openvlthinker: Complex vision-language reasoning via iterative sft-rl cycles},
  author={Deng, Yihe and Bansal, Hritik and Yin, Fan and Peng, Nanyun and Wang, Wei and Chang, Kai-Wei},
  booktitle={The 5th Workshop on Mathematical Reasoning and AI at NeurIPS 2025},
  year={2025},
  url={https://arxiv.org/abs/2503.17352}
}

@inproceedings{zhang2024mathverse,
  title={Mathverse: Does your multi-modal llm truly see the diagrams in visual math problems?},
  author={Zhang, Renrui and Jiang, Dongzhi and Zhang, Yichi and Lin, Haokun and Guo, Ziyu and Qiu, Pengshuo and Zhou, Aojun and Lu, Pan and Chang, Kai-Wei and Qiao, Yu and others},
  booktitle={European Conference on Computer Vision},
  pages={169--186},
  year={2024},
  organization={Springer},
  url={https://arxiv.org/abs/2403.14624}
}

@article{wang2024mathvision,
  title={Measuring multimodal mathematical reasoning with math-vision dataset},
  author={Wang, Ke and Pan, Junting and Shi, Weikang and Lu, Zimu and Ren, Houxing and Zhou, Aojun and Zhan, Mingjie and Li, Hongsheng},
  journal={Advances in Neural Information Processing Systems},
  volume={37},
  pages={95095--95169},
  year={2024},
  url={https://arxiv.org/abs/2402.14804}
}

@article{xiao2024logicvista,
  title={Logicvista: Multimodal llm logical reasoning benchmark in visual contexts},
  author={Xiao, Yijia and Sun, Edward and Liu, Tianyu and Wang, Wei},
  journal={arXiv preprint arXiv:2407.04973},
  year={2024},
  url={https://arxiv.org/abs/2407.04973}
}

@inproceedings{yue2024mmmu,
  title={Mmmu: A massive multi-discipline multimodal understanding and reasoning benchmark for expert agi},
  author={Yue, Xiang and Ni, Yuansheng and Zhang, Kai and Zheng, Tianyu and Liu, Ruoqi and Zhang, Ge and Stevens, Samuel and Jiang, Dongfu and Ren, Weiming and Sun, Yuxuan and others},
  booktitle={Proceedings of the IEEE/CVF Conference on Computer Vision and Pattern Recognition},
  pages={9556--9567},
  year={2024},
  url={https://arxiv.org/abs/2311.16502}
}

@article{chen2024mmstar,
  title={Are we on the right way for evaluating large vision-language models?},
  author={Chen, Lin and Li, Jinsong and Dong, Xiaoyi and Zhang, Pan and Zang, Yuhang and Chen, Zehui and Duan, Haodong and Wang, Jiaqi and Qiao, Yu and Lin, Dahua and others},
  journal={Advances in Neural Information Processing Systems},
  volume={37},
  pages={27056--27087},
  year={2024},
  url={https://arxiv.org/abs/2403.20330}
}

@inproceedings{lu2024mathvista,
  title={MathVista: Evaluating Mathematical Reasoning of Foundation Models in Visual Contexts},
  author={Lu, Pan and Bansal, Hritik and Xia, Tony and Liu, Jiacheng and Li, Chunyuan and Hajishirzi, Hannaneh and Cheng, Hao and Chang, Kai-Wei and Galley, Michel and Gao, Jianfeng},
  booktitle={International Conference on Learning Representations (ICLR)},
  year={2024},
  url={https://arxiv.org/abs/2310.02255}
}

@article{sheng2024hybridflow,
  title   = {HybridFlow: A Flexible and Efficient RLHF Framework},
  author  = {Guangming Sheng and Chi Zhang and Zilingfeng Ye and Xibin Wu and Wang Zhang and Ru Zhang and Yanghua Peng and Haibin Lin and Chuan Wu},
  year    = {2024},
  journal = {arXiv preprint arXiv: 2409.19256},
  url = {https://arxiv.org/abs/2409.19256}
}

@inproceedings{duan2024vlmevalkit,
  title={Vlmevalkit: An open-source toolkit for evaluating large multi-modality models},
  author={Duan, Haodong and Yang, Junming and Qiao, Yuxuan and Fang, Xinyu and Chen, Lin and Liu, Yuan and Dong, Xiaoyi and Zang, Yuhang and Zhang, Pan and Wang, Jiaqi and others},
  booktitle={Proceedings of the 32nd ACM International Conference on Multimedia},
  pages={11198--11201},
  year={2024},
  url = {https://dl.acm.org/doi/abs/10.1145/3664647.3685520}
}

@inproceedings{dong2025stp,
title={{STP}: Self-play {LLM} Theorem Provers with Iterative Conjecturing and Proving},
author={Kefan Dong and Tengyu Ma},
booktitle={Forty-second International Conference on Machine Learning},
year={2025},
url={https://icml.cc/virtual/2025/poster/43472}
}

@article{kuba2025language,
  title={Language Self-Play For Data-Free Training},
  author={Kuba, Jakub Grudzien and Gu, Mengting and Ma, Qi and Tian, Yuandong and Mohan, Vijai},
  journal={arXiv:2509.07414},
  year={2025},
  url={https://arxiv.org/abs/2509.07414}
}

@article{wang2025vision,
  title={Vision-Zero: Scalable VLM Self-Improvement via Strategic Gamified Self-Play},
  author={Wang, Qinsi and Liu, Bo and Zhou, Tianyi and Shi, Jing and Lin, Yueqian and Chen, Yiran and Li, Hai Helen and Wan, Kun and Zhao, Wentian},
  journal={arXiv:2509.25541},
  year={2025},
  url={https://arxiv.org/abs/2509.25541}
}

@article{zhao2025absolute,
  title={Absolute Zero: Reinforced Self-play Reasoning with Zero Data},
  author={Zhao, Andrew and Wu, Yiran and Yue, Yang and Wu, Tong and Xu, Quentin and Lin, Matthieu and Wang, Shenzhi and Wu, Qingyun and Zheng, Zilong and Huang, Gao},
  journal={arXiv:2505.03335},
  year={2025},
  url={https://arxiv.org/abs/2505.03335}
}

@article{liu2025spiral,
  title={SPIRAL: Self-Play on Zero-Sum Games Incentivizes Reasoning via Multi-Agent Multi-Turn Reinforcement Learning},
  author={Liu, Bo and Guertler, Leon and Yu, Simon and Liu, Zichen and Qi, Penghui and Balcells, Daniel and Liu, Mickel and Tan, Cheston and Shi, Weiyan and Lin, Min and others},
  journal={arXiv:2506.24119},
  year={2025},
  url={https://arxiv.org/abs/2506.24119}
}

@article{chen2025spc,
  title={{SPC:} Evolving Self-Play Critic via Adversarial Games for {LLM} Reasoning},
  author={Chen, Jiaqi and Zhang, Bang and Ma, Ruotian and Wang, Peisong and Liang, Xiaodan and Tu, Zhaopeng and Li, Xiaolong and Wong, Kwan-Yee K},
  journal={arXiv:2504.19162},
  year={2025},
  url={https://arxiv.org/abs/2504.19162}
}

@article{huang2025r,
  title={R-Zero: Self-Evolving Reasoning {LLM} from Zero Data},
  author={Huang, Chengsong and Yu, Wenhao and Wang, Xiaoyang and Zhang, Hongming and Li, Zongxia and Li, Ruosen and Huang, Jiaxin and Mi, Haitao and Yu, Dong},
  journal={arXiv:2508.05004},
  year={2025},
  url={https://arxiv.org/abs/2508.05004}
}

@inproceedings{sun2025mm,
  author       = {Linzhuang Sun and
                  Hao Liang and
                  Jingxuan Wei and
                  Bihui Yu and
                  Tianpeng Li and
                  Fan Yang and
                  Zenan Zhou and
                  Wentao Zhang},
  editor       = {Wanxiang Che and
                  Joyce Nabende and
                  Ekaterina Shutova and
                  Mohammad Taher Pilehvar},
  title        = {MM-Verify: Enhancing Multimodal Reasoning with Chain-of-Thought Verification},
  booktitle    = {Proceedings of the 63rd Annual Meeting of the Association for Computational
                  Linguistics (Volume 1: Long Papers), {ACL} 2025, Vienna, Austria,
                  July 27 - August 1, 2025},
  pages        = {14100--14115},
  publisher    = {Association for Computational Linguistics},
  year         = {2025},
  url          = {https://aclanthology.org/2025.acl-long.689/},
  bibsource    = {dblp computer science bibliography, https://dblp.org}
}

@article{wang2025sota,
  title ={SoTA with Less: MCTS-Guided Sample Selection for Data-Efficient Visual
                  Reasoning Self-Improvement},  
  author={Wang, Xiyao and Yang, Zhengyuan and Feng, Chao and Lu, Hongjin and Li, Linjie and Lin, Chung-Ching and Lin, Kevin and Huang, Furong and Wang, Lijuan},
  journal={arXiv:2504.07934},
  year={2025},
  url={https://arxiv.org/abs/2504.07934}
}

@article{zhang2025r1,
  title= {{R1-VL:} Learning to Reason with Multimodal Large Language Models
                  via Step-wise Group Relative Policy Optimization},  
  author={Zhang, Jingyi and Huang, Jiaxing and Yao, Huanjin and Liu, Shunyu and Zhang, Xikun and Lu, Shijian and Tao, Dacheng},
  journal={arXiv:2503.12937},
  year={2025},
  url={https://arxiv.org/abs/2503.12937}
}

@inproceedings{
liu2024diving,
title={Diving into Self-Evolving Training for Multimodal Reasoning},
author={Wei Liu and Junlong Li and Xiwen Zhang and Fan Zhou and Yu Cheng and Junxian He},
booktitle={Forty-second International Conference on Machine Learning},
year={2025},
url={https://icml.cc/virtual/2025/poster/44984
}
}

@article{zhao2025genprm,
  title        = {GenPRM: Scaling Test-Time Compute of Process Reward Models via Generative
                  Reasoning},  
  author={Zhao, Jian and Liu, Runze and Zhang, Kaiyan and Zhou, Zhimu and Gao, Junqi and Li, Dong and Lyu, Jiafei and Qian, Zhouyi and Qi, Biqing and Li, Xiu and others},
  journal={arXiv:2504.00891},
  year={2025},
  url={https://arxiv.org/abs/2504.00891}
}

@article{yao2024mulberry,
  title= {Mulberry: Empowering {MLLM} with o1-like Reasoning and Reflection via Collective Monte Carlo Tree Search},
  author={Yao, Huanjin and Huang, Jiaxing and Wu, Wenhao and Zhang, Jingyi and Wang, Yibo and Liu, Shunyu and Wang, Yingjie and Song, Yuxin and Feng, Haocheng and Shen, Li and others},
  journal={arXiv:2412.18319},
  year={2024},
  url={https://arxiv.org/abs/2412.18319}
}

@article{yang2025r1,
  title={R1-Onevision: Advancing Generalized Multimodal Reasoning through Cross-Modal
                  Formalization},
  author={Yang, Yi and He, Xiaoxuan and Pan, Hongkun and Jiang, Xiyan and Deng, Yan and Yang, Xingtao and Lu, Haoyu and Yin, Dacheng and Rao, Fengyun and Zhu, Minfeng and others},
  journal={arXiv:2503.10615},
  year={2025},
  url={https://arxiv.org/abs/2503.10615}
}

@article{huang2025vision,
  title        = {Vision-R1: Incentivizing Reasoning Capability in Multimodal Large
                  Language Models},  
  author={Huang, Wenxuan and Jia, Bohan and Zhai, Zijie and Cao, Shaosheng and Ye, Zheyu and Zhao, Fei and Xu, Zhe and Hu, Yao and Lin, Shaohui},
  journal={arXiv:2503.06749},
  year={2025},
  url={https://arxiv.org/abs/2503.06749}
}

@article{yu2025perception,
  title        = {Perception-R1: Pioneering Perception Policy with Reinforcement Learning},
  author={Yu, En and Lin, Kangheng and Zhao, Liang and Yin, Jisheng and Wei, Yana and Peng, Yuang and Wei, Haoran and Sun, Jianjian and Han, Chunrui and Ge, Zheng and others},
  journal={arXiv:2504.07954},
  year={2025},
  url={https://arxiv.org/abs/2504.07954}
}

@article{liu2025spice,
  title={{SPICE:} Self-Play In Corpus Environments Improves Reasoning},
  author={Liu, Bo and Jin, Chuanyang and Kim, Seungone and Yuan, Weizhe and Zhao, Wenting and Kulikov, Ilia and Li, Xian and Sukhbaatar, Sainbayar and Lanchantin, Jack and Weston, Jason},
  journal={arXiv preprint arXiv:2510.24684},
  year={2025},
  url={https://arxiv.org/abs/2510.24684}
}

@inproceedings{kwon2023vllm,
  title={Efficient Memory Management for Large Language Model Serving with PagedAttention},
  author={Woosuk Kwon and Zhuohan Li and Siyuan Zhuang and Ying Sheng and Lianmin Zheng and Cody Hao Yu and Joseph E. Gonzalez and Hao Zhang and Ion Stoica},
  booktitle={Proceedings of the ACM SIGOPS 29th Symposium on Operating Systems Principles},
  year={2023},
  url={https://dl.acm.org/doi/10.1145/3600006.3613165}
}
\newpage
\appendix
\section{Appendix}
\label{sec:appendix}
\subsection{Dataset}
\label{sec:appendix:dataset}
The images we use are sourced from two datasets: OpenVLThinker-GRPO-hard\footnote{\url{https://huggingface.co/datasets/ydeng9/OpenVLThinker-grpo-hard}}, which contains approximately 6K geometric images, and OpenVLThinker-GRPO-medium\footnote{\url{https://huggingface.co/datasets/ydeng9/OpenVLThinker-grpo-medium}}, which comprises around 3K images of maps, charts, and object spatial relationships. From these, we select 4,000 images and train only on the raw, unlabeled image data.
\subsection{Training Settings}
\label{sec:appendix:training}
V-Zero framework is based on \textbf{verl} codebase. We employ vLLM~\citep{kwon2023vllm} as the rollout engine during training, the inference temperature is set to 1.0 across all processes. The The maximum number of tokens for the response is set to 2048 for Questioner training and 4096 for Solver training. We use a KL divergence coefficient of 0.01 to constrain policy drift. To manage memory efficiency, we employ Fully Sharded Data Parallel (FSDP) with parameter offloading. A single V-Zero iteration requires about 9 hours to complete.

\subsection{Evaluation Details}
\label{sec:appendix:evaluation}
We conduct all evaluations using the \textbf{VLMEvalKit} codebase, a standardized open-source framework for VLM assessment. We use the default generation temperature 0.0 (greedy decoding) to eliminate randomness and maximize the determinism of the results. The maximum generation length of token is set to 2,048.

For the answer extraction and matching process (i.e., the "Judge" role), we employ Qwen3-VL-8B-Instruct. We selected this model as the evaluator because the task involves primarily straightforward answer parsing and option matching, for which a lightweight yet capable 8B model offers an optimal trade-off between accuracy and evaluation efficiency.

\subsection{Prompts}




\begin{tcolorbox}[breakable, colback=white, colframe=Maroon, title={Prompt for Question Generation}]
\textbf{System Prompt:}\\
You are a professional question designer.

\textbf{User Prompt:}\\
Create a multiple-choice question based on the image. Let's think step by step.
First, you must fully perceive the image, extracting any valuable visual information from it and generate a detailed visual description of the image.\\
Then, write a multiple-choice question that includes necessary conditions. \\
- Make sure the question provides sufficient information to be answered. Use phrases like "If..." or "Given that..." to state condition shown in visual description if it's a geometry question.\\
- The question must include four options, one of which is the correct answer. \\
- Provide the correct answer to the generated question without thinking. It must be one of A/B/C/D, and MUST BE enclosed within <answer> </answer> tags.\\
- Any question type other than multiple-choice is STRICTLY FORBIDDEN!\\
\\
Your MUST response in this format:\\
\\
<description>\\
\{Visual description you extract from the image\}\\
</description>\\

<question>\\
\{Write a complete multiple-choice question that states all necessary conditions clearly, followed by exactly 4 answer options A B C D\}\\
</question>\\

<answer>\\
\{Correct answer option A/B/C/D\} \\
</answer>\\
\\
DO NOT output anything else—no explanations, no extra markup.\\
\end{tcolorbox}

\begin{tcolorbox}[breakable, colback=white, colframe=Maroon, title={Prompt for Answer Generation}]
\textbf{System Prompt:}\\
You are a professional question solver.

\textbf{User Prompt:}\\
Solve the multiple-choice question based on the provided image. Let's think step by step.\\
First, you must fully perceive the image, extracting any valuable visual information from it.\\
Then, solve the question, give the correct choice option.\\

The reasoning process MUST BE enclosed within <think> </think> tags.\\
The final answer MUST BE single option A/B/C/D put in \verb|\boxed{}|.
\end{tcolorbox}
\begin{tcolorbox}[breakable, colback=white, colframe=Maroon, title={Prompt for Question Difficulty Evaluation}]
\textbf{System Prompt:}\\
You are a professional question evaluator.

\textbf{User Prompt:}\\
Score the difficulty of the following visual reasoning question on a scale from 0.0 (easiest) to 1.0 (hardest), where:\\
\\
0.1–0.4 (Simple): Basic object counting or simple chart reading;\\
0.4–0.7 (Medium): Application of geometric relations or chart trend analysis;\\
0.7–1.0 (Hard): Complex geometric proofs or multi-step algebraic reasoning.\\
\\
Output only a single number between 0.0 and 1.0. Do not include any other text, explanation, or formatting.
\end{tcolorbox}

\end{document}